\documentclass[runningheads,a4paper]{llncs}

\usepackage{amssymb}
\usepackage{graphicx}
\usepackage{listings}

\usepackage{url}
\urldef{\mailsa}\path|piek.vossen@vu.nl,|   
\urldef{\mailsb}\path|a.k.cybulska@vu.nl|   
\newcommand{\keywords}[1]{\par\addvspace\baselineskip
\noindent\keywordname\enspace\ignorespaces#1}

\begin{document}

\mainmatter  

\title{Identity and Granularity of Events in Text}

\titlerunning{Identity and granularity of events}

%
%
\author{Piek Vossen%
\and Agata Cybulska}
\authorrunning{Vossen and Cybulska: Identity and Granularity of Events in Text}

\institute{Computational Lexicology and Terminology Lab, Faculty of Humanities,\\ VU University Amsterdam,\\
De Boelelaan 1105, 1081HV Amsterdam, Netherlands\\
\mailsa\\
\mailsb\\
\url{http://cltl.nl}}

%
%


\maketitle

\begin{abstract}
In this paper we describe a method to detect event descriptions in different news articles and to model the semantics of events and their components using RDF representations. We compare these descriptions to solve a cross-document event coreference task. Our component approach to event semantics defines identity and granularity of events at different levels. It performs close to state-of-the-art approaches on the cross-document event coreference task, while outperforming other works when assuming similar quality of event detection. We demonstrate how granularity and identity are interconnected and we discuss how semantic anomaly could be used to define differences between coreference, subevent and topical relations.
\keywords{Event coreference, event identity, event relations}
\end{abstract}

\section{Introduction}
News and blogs are common media that report on events that took place in the world. In the case of events with impact, we can expect that many different sources discuss the same event, partially providing the same information and partially differing from each other either in terms of the facts or perspective on these facts. Collections of news and blogs around the same topic therefore represent a challenging and natural task for cross-document event coreference. If done properly, event coreference resolution can be used to link event data across many different sources, resulting in deduplication and aggregation of data around events but also showing the different perspectives of these sources \cite{2016kbs}. 

The task of cross-document event coreference is however far from trivial. Events exist within their temporal boundaries. The same type of event involving the same participants and the same action at a different point of time is not the same event, e.g. \textit{John gave Mary the book on Tuesday} is a different event from \textit{John gave Mary the book on Wednesday}. On the other hand, \textit{Mary gave John the book on Tuesday} is also a different event, even though all event components are the same as for the first event but the roles differ. The precise semantics of these descriptions can be used to define identity across events. This becomes more complicated when we underspecify and quantify the information. In the case of \textit{John gave Mary several books this week} or \textit{Mary often gets books from people} the precise details are not given and matching with the previous descriptions becomes more difficult. Also, it is not clear if the quantification should be seen as a quantification of the object \textit{books} or of the event of \textit{giving}. Furthermore, the action itself can be described in many different ways (\textit{gets/takes/receives/borrows/buys/obtains}), exhibiting different manners or perspectives on the same event.

These examples clearly show that establishing identity across event descriptions is a hard problem to solve and on the one hand involves semantic components that play a role but on the other hand requires a robust matching across these components. In \cite{CYBULSKA:2013}, we therefore described a model to measure identity across events as a function of the similarity of the event components. Such a model can be optimized on an annotated data set to weigh the contribution of the components for establishing event identity.

When implementing such a model for extracting and identifying events from raw text, another problem arises. The components necessary to compare events are spread over a complete article and are hardly found in a single sentence. Sentence-based approaches to event coreference cannot deal with the fact that event components are mentioned at different places of the text. In \cite{CYBULSKA:2015cicling} a bag of events approach is suggested to group event components per source article and compare these across different articles. However, the implementation of the bag of events approach described in \cite{CYBULSKA:2015cicling} was only tested on true mentions and did not consider the extraction of events from text. Furthermore, in the data set used in the experiments reported in  \cite{CYBULSKA:2015cicling} only a few sentences per news article are annotated, which means that only little information could be aggregated across these sentences.

In this paper, we describe a new implementation of the bag of events model proposed in \cite{CYBULSKA:2015cicling} which completely processes news articles starting from raw text. First event instance representations are build from all mentions throughout the text within a single article and next these representations are compared across articles. The method employed here makes a distinction between mentions within an article and across articles on the one hand and event instances that are stable across these mentions on the other hand \cite{FOKKENS:2014isa}. Likewise, we can compare event mentions but also event instance representations. Our system can be highly parameterized to establish identity. This can result in different levels of granularity of determined event instances. More loose constraints result in extreme aggregation and lumping, whereas very strict definitions result in each mention to be unique. We show that our instance-based approach performs close to state-of-the-art machine learning approaches and it out-performs these approaches when a similar quality of event detection is assumed.

In section \ref{sec:related-work}, we summarize previous work that is related to this task. In section \ref{sec:data}, we describe the event model and data set on which we evaluate our approach.
Section \ref{sec:gold-data} describes the bag of events approach that we proposed previously, which collects event information per document so that it can be used to solve event coreference. We present earlier implementation examples of the approach and their evaluation results achieved on true mentions in section \ref{sec:evaluation-true}. Next in section \ref{sec:implementation} 
we propose a new, door-to-door implementation 
of the bag of events approach. We evaluate the implementation and compare results with the state-of-the-art in section \ref{sec:evaluation}. In section \ref{sec:discussion}, we discuss the results and speculate on the notion of granularity in relation to identity. We conclude in section \ref{sec:conclusion}.

\section{Related work}
\label{sec:related-work}
Event coreference resolution is a difficult task of determining whether two event mentions refer to the same event instance. After the original focus on entity coreference, recently there is more and more interest in the field in event coreference resolution.

Some supervised approaches to event coreference resolution have been proposed by Humphreys et al. \cite{humpreys}, Bagga and Baldwin \cite{bagga_baldwin_event_coref}, Ahn \cite{david_ahn_event_extr}, or Chen and Ji \cite{chen-graph}. These methods rely on supervised learning of rich linguistics features to determine coreference in a pairwise model and are strongly domain dependent. This is not the case for the method presented in section \ref{sec:evaluation} which is unsupervised. Furthermore, these works depend on local decisions without consideration of global event distribution at document level. The bag of events model implemented here allows us to overcome this deficiency.  

More recently there are a few more works also using rich linguistic features to solve event coreference. One of them is the unsupervised approach of Bejan and Harabagiu \cite{Bejan-Harabagiu-2010} which relies on lexical, POS, event class and WordNet features as well as feature combinations. Another approach is the one of Liu et al. \cite{liu} and the most recent feature-rich model of Yang et al. \cite{YangCF15}. Our method is similar to the approach of \cite{liu} in that it facilitates propagation of information about event participants between event mentions but our heuristic does this more globally by employing an instance level of event representation that aggregates this information at the document level as is also done \cite{YangCF15}. We differ from \cite{YangCF15} in that we use an RDF representation to do a logical comparison, whereas they do clustering using document level features. Our approach also differs from another recent approach to event coreference by Lee et al. \cite{Lee-etal-2012}, by making a distinction between specific entity types, whereas \cite{Lee-etal-2012} disregard entity type information.

\section{Event model and data}
\label{sec:data}

We model events as a combination of five slots. These five slots correspond to different elements of event information such as the action slot (or event trigger following the ACE (\cite{ACE-event-2005}) terminology) and four kinds of event arguments: time, location, human and non-human participant slots (see \cite{CYBULSKA2014nwr1}).
The next quote shows an excerpt from topic one, text number seven of the ECB corpus (\cite{Bejan-Harabagiu-2010}).
\begin{quotation}
\textit{The ``American Pie" actress has entered Promises for undisclosed reasons. The actress, 33, reportedly headed to a Malibu treatment facility on Tuesday.}
\end{quotation}
Consider two event templates presenting the distribution of event information over the five event slots in the two example sentences (tables \ref{tab:ecb-1} and \ref{tab:ecb-2}).
\begin{table}[htp]
\caption{Sentence template ECB topic 1, text 7, sentence 1}
{\scriptsize 
\begin{center}
\begin{tabular}{|l|l|}
\hline
Action & \textit{entered}\\
Time &  N/A\\
Location &  \textit{Promises}\\
Human Participant & \textit{actress}\\
Non-Human Participant & N/A\\
\hline
\end{tabular}
\end{center}
}

\label{tab:ecb-1}
\end{table}%
\begin{table}[htp]
\caption{Sentence template ECB topic 1, text 7, sentence 2}
{\scriptsize 
\begin{center}
\begin{tabular}{|l|l|}
\hline
Action & \textit{headed}\\
Time & \textit{on Tuesday}\\
Location & \textit{to a Malibu treatment facility}\\
Human Participant & \textit{actress}\\
Non-Human Participant&  N/A\\
\hline
\end{tabular}
\end{center}
}
\label{tab:ecb-2}
\end{table}%
This event model has been employed to annotate the ECB+ data set (\cite{CYBULSKA+VOSSEN:2014}) which is used in the experiments with event coreference described in the following sections.
The approach to event coreference used in this work determines coreference between mentions of events through compatibility of slots of the five slot template. 
If event mentions are coreferent, they refer to the same event instance. Two or more event mentions are coreferent if they describe actions that happen or hold true in the same time and place and with involvement of the same participants.

The EventCorefBank (ECB, \cite{Bejan-Harabagiu-2010}) was developed to test cross-document event coreference resolution. It consists of 43 different topics which correspond to seminal events, each topic with about 10 to 20 news articles reporting on a seminal event. Across articles from a topic, mentions of events are coreferent. The ECB+ corpus \cite{CYBULSKA+VOSSEN:2014} is an extended and re-annotated version of the ECB. To each of the 43 ECB topics new texts were added about different event instances of the same event type. For example in addition to the topic of a particular celebrity checking into rehab covered in the ECB, to  ECB+ descriptions were added of another event instance involving a different celebrity checking into another rehab facility. Likewise, the authors of ECB+ increased the referential ambiguity of event mentions. Table~\ref{fig:ecb+topics} shows some examples of seminal events represented in ECB+ with two different event instances per topic. Table~\ref{tab:ecb-stats} shows some statistics on the data set, most notably there are 6833 mentions of events annotated and
1958 coreference chains (instances). On average, 1.8 sentence per article was annotated for experiments on event coreference.

\begin{table}[t]
{\tiny
  \centering
\caption{\label{fig:ecb+topics} Overview of seminal events in ECB and ECB+, topics 1-10}
    \begin{tabular}{|r|r|r|r|r|r|r|r|r|r|}
    \hline
    Topic & Seminal Event type & \multicolumn{2}{c|}{Human participant} & \multicolumn{2}{c|}{Time} & \multicolumn{2}{c|}{Location} & \multicolumn{2}{c|}{Nr of docs} \\
          &       & ECB   & ECB+  & ECB   & ECB+  & ECB   & ECB+  & ECB   & ECB+ \\
    \hline
    1     & rehab check-in & T. Reid & L. Lohan & 2008  & 2013  & Malibu & R. Mirage & 18    & 21 \\
    2     & Oscars host announced & H. Jackman & E. Degeneres & 2010  & 2014  & -     & -     & 10    & 11 \\
    3     & inmate escape & Brian Nicols & A.J. Corneaux & 2008  & 2009  & court house & prison & 9     & 11 \\
	   & &  4 dead & Jr & & & Atlanta & Texas & & \\ 
    4     & death & B. Page & E. Williams & 2008  & 2013  & LA    & LA    & 14    & 10 \\
    5     & head  & Philadelphia & Philadelphia & 2008  & 2005  & -     & -     & 13    & 10 \\
         &  coach & 76ers & 76ers &  &   &      &      &   & \\

	   & fired & M. Cheeks & J. O'Brien & & & & & & \\ 
    6     & "Hunger Games"  & C. Weitz & G. Ross & 2008  & 2012  & -     & -     & 9     & 11 \\    
	   & sequel negotiations &  & & & & & & & \\ 
    7     & IBF, IBO, WBO & W. Klitchko  & W. Klitchko & 2008  & 2012  & Germany & Switzerland & 11    & 11 \\    
	   &  titles defended & H. Rahman & T. Thompson & & & & & & \\ 
    8     & explosion at bank & -     & -     & 2008  & 2012  & Oregon & Athens & 8     & 11 \\
    9     & ESA changes & Bush  & Obama & 2008  & 2009  & -     & -     & 10    & 13 \\
    10    & eigth-year offer & Angels  & Red Socks & 2008  & 2008  & -     & -     & 8     & 13 \\    
	   &  & M. Teixeira & M. Teixeira & & & & & &\\  
    45    & murder & S. Peterson  & C. Simpson &   & 2012  & -     & -     & 8     & 12 \\    
	   &  & L. Peterson & K. Flynn & & & & & &\\    \hline
    \end{tabular}%
  }
\end{table}

\begin{table}[htp]
\caption{ECB+ statistics }
{\footnotesize 
\begin{center}
\begin{tabular}{|l|r|}
\hline
ECB+&\#\\
\hline
Topics& 43 \\
Texts & 982 \\
Action mentions & 6833 \\
Location mentions & 1173 \\
Time mentions &1093 \\
Human participant mentions & 4615 \\
Non-human participant mentions &1408 \\
Coreference chains &1958 \\
\hline
\end{tabular}
\end{center}
}
\label{tab:ecb-stats}
\end{table}%

%










\section{Modelling event coreference: bag of events approach}
\label{sec:gold-data}

It is pretty much common practice to use information coming from event arguments for event coreference resolution (\cite{humpreys}, \cite{chen-graph}, \cite{chen-coref-alg}, \cite{chen-etal-2011}, \cite{Bejan-Harabagiu-2010}, \cite{Lee-etal-2012}, \cite{CYBULSKA:2013}, \cite{liu} among others). But using entities for event coreference resolution is complicated by the fact that event descriptions within a sentence often lack pieces of information. As pointed out by \cite{humpreys} it could be the case however that a lacking piece of information might be available elsewhere within discourse borders. This is a challenge for pairwise models comparing separate event mentions with one another on the sentence level. To be able to fully make use of information coming from event arguments, instead of looking at event information available within the same sentence, we propose to take a broader look at event descriptions surrounding the event mention in question within a unit of discourse. In this study we consider a document (a news article) as the unit of discourse.

The bag of events approach (for details see \cite{CYBULSKA:2015cicling}) translates the structure of event descriptions into event templates for event coreference resolution. An event template can be created on different levels of information, such as a sentence, a paragraph or an entire document. The approach explicitly employs  discourse structure to account for challenges following from the uneven distribution of event information across sentences of a document.
Two templates are filled: a sentence and a document template. A sentence template collects event information from the sentence of an active action mention (tables \ref{tab:ecb-1} and \ref{tab:ecb-2}). By filling in a document template, one creates a ``bag of events" for a document, that could be seen as a kind of document ``summary" (table \ref{tab:event-sum}).
\begin{table}[htp]
\caption{Document template ECB topic 1, text 7, sentences 1-2}
{\scriptsize 
\begin{center}
\begin{tabular}{|l|l|}
\hline
Action & \textit{entered, headed} \\
Time &  \textit{on Tuesday} \\
Location & \textit{Promises, to a Malibu treatment facility} \\
Human Participant & \textit{actress}\\
Non-Human Participant& N/A \\
\hline
\end{tabular}
\end{center}
}
\label{tab:event-sum}
\end{table}

This heuristic employs clues coming from discourse structure and namely those implied by discourse borders. 
Descriptions of different event mentions occurring within a discourse unit, whether coreferent or related in some other way, unless stated otherwise, tend to share elements of their context.
In our example text fragment the first sentence reveals that an actress has entered a rehab facility. From the second sentence the reader finds out where the facility is located (Malibu) and when the ``American Pie" actress headed to the treatment center. It is clear to the reader of the example text fragment from the quotation that both events described in sentence one and two, happened on Tuesday. Also both sentences mention the same rehab center in Malibu. These observations are crucial for the approach.

The bag of events method can be implemented in different ways. In the following section \ref{sec:evaluation-true} we will briefly look at experiments with implementations as one- and as two-step classification tasks (the two-step implementation is described in detail in \cite{CYBULSKA:2015cicling}). This implementation uses true mentions of event data and represents the document-based bag of events features as a loose set. In the one-step approach document-based bag of events features are combined with sentence-based features for pairwise mention comparison. In the two-step approach the bag of events features are first used for document clustering and the sentence-based features are used for pairwise mention comparison within a cluster. In section \ref{sec:implementation}, we describe another two-step implementation that extracts all event and entity data from the text and represents events as RDF instances, which aggregate all event data across the coreferential mentions in a single news article. These document-based RDF representations are  more specific than the bag of events representations. In the second step, these RDF representations are compared semantically across all documents within a topic. The final instance results are mapped back to all the mentions for evaluation. We evaluate the RDF implementation in section \ref{sec:evaluation}.

\section{Experiments with coreference on true mentions} 
\label{sec:evaluation-true}

\subsection{One- vs. two-step classification}

The specifics of the implementation of the two-step approach can be found in \cite{CYBULSKA:2015cicling}.

The two-step bag of events approach starts with filling in a document template, accumulating mentions of the five event slots: actions, locations, times, human and non-human participants from a document, as exemplified in table \ref{tab:event-sum}.
In a document template there is no distinction made between pieces of event information coming from different sentences of a document and no information is kept about elements being part of different mentions. A document template can be seen as a bag of events and event arguments. The template stores unique lemmas, to be precise a set of unique lemmas per event template slot.
Document-based features from the template are used for preliminary document clustering. A supervised decision tree classifier (hereafter \textit{DT}) determines whether two document templates share corefering event mentions. After all unique pairs of document templates from the test set have been classified by means of the DT classifier, ``compatible" pairs are merged into document clusters based on pair overlap.

In the second step of the approach coreference is solved in a pairwise model between action mentions per document cluster created in step 1. For this task sentence templates are filled  per true action mention from the corpus. Sentence templates collect event information from the sentence (consider examples in table \ref{tab:ecb-1} and \ref{tab:ecb-2}). Pairs of sentence templates translate into features indicating compatibility across five template slots. A supervised classifier solves coreference between all unique pairs of action mentions per document cluster and finally pairs sharing common mentions are chained into equivalence classes.

In the one-step implementation of the approach all possible unique pairs of action mentions from the corpus are used as the starting point for classification. No initial document clustering is performed.
For every action mention a sentence template is filled (see examples in table \ref{tab:ecb-1} and \ref{tab:ecb-2}). Also, for every corpus document a document template is filled. Bag of events features indicating the degree of overlap between documents, from which two active mentions come from, are used for classification. In the one-step approach document features are used by a classifier together with sentence-based features; these combined create a feature vector per active action pair. One DT classifier is used to determine event coreference in a pairwise model. Pairs of mentions are classified based on a mix of information from a sentence and from a document. Finally, corefering pairs with overlap are merged into equivalence classes.

The one-step classification is implementation-wise simpler but it is computationally much more expensive. Ultimately every action mention has to be compared with every other action mention from the data set. This is a drawback of the one-step approach. On the other hand, it could be of advantage to have different types of information (sentence- and document-based) available simultaneously to determine event mention coreference.

\subsection{Experiment set-up}

For coreference experiments on true mentions we used a subset of ECB+ annotations (based on a list of 1840 selected sentences, \cite{CYBULSKA+VOSSEN:2014}), that were additionally reviewed with focus on coreference relations. We divided the corpus into a training set (topics 1-35) and test set (topics 36-45). 

The ECB+ texts are available in the XML format. The texts are tokenized, hence no sentence segmentation nor tokenization needed to be done. We POS-tagged (for the purpose of proper verb lemmatization) and lemmatized the corpus sentences. We used tools from the Natural Language Toolkit (\cite{nltk:2009}, NLTK version 2.0.4): the NLTK's default POS tagger, Word-Net lemmatizer\footnote{\url{www.nltk.org/ modules/nltk/stem/wordnet.html}} as well as WordNet synset assignment by the NLTK\footnote{http://nltk.org/\_modules/nltk/corpus/reader/wordnet.html}. For machine learning experiments we used scikit-learn (\cite{scikitlearn}). 

 \begin{table*}[htbp]
  \caption{\label{font-table} Features grouped into four categories: L-Lemma based, A-Action similarity, D-location within Discourse, E-Entity coreference and S-Synset based.}
 \begin{center}
 \centering 
 \begin{tabular}{|l|l |l| l| l|}
       \hline
      \multicolumn{2}{|l|}{Event Slot }&Mentions&Feature Kind&Explanation \\
       \hline
      Action & \multicolumn{2}{|l|}{Active}& Lemma overlap (L)& Numeric feature: overlap \%.\\      
      &\multicolumn{2}{|l|}{mentions} &Synset overlap (S)& Numeric: overlap \%. \\
       &\multicolumn{2}{|l|}{} &Action similarity (A)& Numeric: Leacock and Chodorow.\\ 
      &\multicolumn{2}{|l|}{} & Discourse location (D)& Binary: \\ 
      &\multicolumn{2}{|l|}{} &- document & - the same document or not.\\
      &\multicolumn{2}{|l|}{} &- sentence& - the same sentence or not.\\
       \cline{2-5}
       & \multicolumn{2}{|l|}{Sent. or doc.}& Lemma overlap (L)&Numeric: overlap \%.\\ 
       &\multicolumn{2}{|l|}{mentions} &Synset overlap  (S)& Numeric: overlap \%. \\
       \hline
       Location&\multicolumn{2}{|l|}{Sent. or doc} & Lemma overlap (L)&Numeric: overlap \%.\\
       & \multicolumn{2}{|l|}{mentions} &Entity coreference (E)& Numeric: cosine similarity. \\
        &\multicolumn{2}{|l|}{} &Synset overlap (S)& Numeric: overlap \%. \\
       \hline
       Time&\multicolumn{2}{|l|}{Sent. or doc } & Lemma overlap (L)&Numeric: overlap \%.\\
        &  \multicolumn{2}{|l|}{mentions} &Entity coreference (E)& Numeric: cosine similarity.\\
        &\multicolumn{2}{|l|}{} &Synset overlap (S)& Numeric: overlap \%. \\
\hline
      Human&\multicolumn{2}{|l|}{Sent. or doc}&Lemma overlap (L)& Numeric: overlap \%.\\
        Participant&  \multicolumn{2}{|l|}{mentions} &Entity coreference (E)& Numeric: cosine similarity.\\
       & \multicolumn{2}{|l|}{} &Synset overlap (S)& Numeric: overlap \%. \\
\hline
       Non-&\multicolumn{2}{|l|}{Sent. or doc}& Lemma overlap (L)& Numeric: overlap \%.\\
       Human& \multicolumn{2}{|l|}{mentions} &Entity coreference (E)& Numeric: cosine similarity.\\
       Participant&\multicolumn{2}{|l|}{} &Synset overlap (S)& Numeric: overlap \%. \\
       \hline
 \end{tabular}
 \end{center}
 \label{tab:features}
 \end{table*}

In the experiments different features were assigned values per event slot (see Table \ref{tab:features}). 
Note, that frequently one ends up with multiple entity mentions from the same sentence for an action mention (the relation between an action and involved entities is not annotated in ECB+).
All entity mentions from the sentence (or a document in case of bag of events features) are considered.
In case of document templates features referring to active action mentions were disregarded, instead action mentions from a document were considered. All feature values were rounded to the first decimal point.

We experimented with a few feature sets, considering per event slot lemma features only (L), or combining them with other features described in Table \ref{tab:features}.
Before fed to a classifier, missing values were imputed (no normalization was needed for the scikit-learn DT algorithm).
All classifiers were trained on an unbalanced number of pairs of examples from the training set. 
We used grid search with ten fold cross-validation to optimize the hyper-parameters (maximum depth, criterion, minimum samples leafs and split) of the decision-tree algorithm.

\subsection{Evaluation on true mentions from ECB+}

We will consider two baselines: a singleton baseline and a rule-based lemma match baseline.
The singleton baseline considers event coreference evaluation scores generated taking into account all action mentions as singletons. 
The rule-based lemma baseline generates event coreference clusters based on full overlap between lemma or lemmas of compared event triggers (action slot) from the test set.
Table \ref{tab:baseline-results} presents baselines' results in terms of recall (R), precision (P) and F-score (F) by employing the following metrics: 
MUC (\cite{Vilain-etal-1995}), B3 (\cite{b3}), CEAF (\cite{Luo-2005}), BLANC (\cite{RecasensHovy-2011}), and CoNLL F1 (\cite{Phradan-etal-2011}). 
 \begin{table*}[htbp]
 {\scriptsize
  \caption{Baseline results on the ECB+: singleton baseline and lemma match of event triggers evaluated in MUC, B3, mention-based CEAF, BLANC and CoNLL F. } 
 \centering
 \begin{center}
 \begin{tabular}{|l |l|c| c| c| c| c| c| c| c| c| c |c|}
       \hline
       Baseline&\multicolumn{3}{c|}{MUC} & \multicolumn{3}{c|}{B3} & CEAF &  \multicolumn{3}{c|}{BLANC} & CoNLL\\
       &R & P & F & R & P & F &R/P/F & R & P & F & F\\
       \hline
Singleton Baseline&0&0&0&45&100&62&45&50&50&50&39\\ 
Action Lemma Baseline&71&60&65&68&58&63&51&65&62&63&\textbf{62}\\ 
\hline
 \end{tabular}
 \end{center}
  \label{tab:baseline-results}
  }
 \end{table*}
 
When discussing event coreference scores must be noted that some of the commonly used metrics depend on the evaluation data set, with scores going up or down with the number of singleton items in the data \cite{RecasensHovy-2011}. 
Our singleton baseline gives us zero scores in MUC, which is understandable due to the fact that the MUC measure promotes longer chains. B3 on the other hand seems to give additional points to responses with more singletons, hence the remarkably high scores achieved by the baseline in B3.
CEAF and BLANC as well as the CoNLL measures (the latter being an average of MUC, B3 and entity CEAF) give more realistic results.
The lemma baseline reaches 62\% CoNLL F1. 

Table \ref{tab:results} evaluates final clusters of coreferent action mentions produced in the experiments by means of the one- and two-step classification when employing different features. 

 \begin{table*}[htbp]
 {\scriptsize
  \caption{Bag of events approach to event coreference resolution, evaluated on the ECB+ in MUC, B3, mention-based CEAF, BLANC and CoNLL F measures. } 
 \centering
 \begin{center}
 \begin{tabular}{|c |c|c|c|c|c|c| c|c| c| c| c| c| c| c| c| c| c|c}
       \hline
       \multicolumn{3}{|c|}{Step1}&\multicolumn{3}{c|}{Step2}&\multicolumn{3}{c|}{MUC} & \multicolumn{3}{c|}{B3} & CEAF &  \multicolumn{3}{c|}{BLANC} & CoNLL\\
       \hline
       Alg&Slot&Fea-&Alg&Slot&Fea-&R & P & F & R & P & F & F & R & P & F & F\\ 
       &Nr&tures&&Nr&tures&& &  &  &  & &  & &  &  &  \\
       \hline 
       -&-&-&DT&5&L&61&76&68&66&79&72&61&67&69&68&70\\
\hline
-&-&-&DT&5&L+docL &65&80&71&68&83&75&64&69&73&71&72\\
\hline 
DT&5&docL&DT&5&L&71&75&73&71&77&74&64&71&71&71&73\\
\hline 
DT&5&docL&DT&5&LDES&71&75&73&71&78&74&64&72&71&72&\textbf{73}\\
DT&2&docL&DT&2&LDES&76&70&73&74&68&71&61&74&68&70&70\\
\hline 
DT&5&docL&DT&5&LADES&71&75&73&71&78&74&64&72&71&72&73\\
\hline
 \end{tabular}
 \end{center}
 \label{tab:results}
 }
 \end{table*} 

When considering bag of events classifiers using exclusively lemma features L (row two and three), the two-step approach reached a 1\% higher CoNLL F-score than the one-step approach with document-based lemma features (\textit{docL}). The one-step method achieved in BLANC a 2\% better precision but a 2\% lower recall. This is understandable. In a two-step implementation when document clusters are created some precision is lost. In a one-step classification specific sentence information is always available for the classifier hence we see slightly higher precision scores (also in other metrics).

The best coreference evaluation scores with the highest CoNLL F-score of 73\% and BLANC F of 72\% were reached by the two-step bag of events approach with a combination of the DT document  classifier using feature set L (document-based hence \textit{docL}) across five event slots and the DT
sentence classifier when employing features LDES
(see Table \ref{tab:features} for a description of features).
Adding action similarity (A) on top of LDES features in step two, does not make any difference on decision tree classifiers with a maximum depth of 5 using five slot templates.
Our best CoNLL F-score  of 73\% is an 11\% improvement over the strong rule based event trigger lemma baseline, and a 34\% increase over the singleton baseline.

To quantify the contribution of document features, we contrast the results of classifiers using bag of events features with scores achieved when disregarding document features.
The results reached with sentence template classification only (without any document features, row one in table \ref{tab:results}), give us some insights into the impact of the document features on our experiment.
Note that one-step classification without preliminary document template clustering is computationally much more expensive than a two-step approach, which ultimately takes into account much less item pairs thanks to the initial document template clustering.
The DT sentence template classifier trained on an unbalanced training set reaches 
70\% CoNLL F. 
This is 8\% better than the strong baseline disregarding event arguments, but only 3\% less than the two-step bag of events approach and 2\% less than the one-step classification with document features. The reason for the relatively small contribution by document features could owe to the fact that in the ECB+ corpus not that many sentences are annotated per text.
1840 sentences are annotated in 982 corpus texts, i.e. 1.87 sentence per text.
We expect that the impact of document features would be bigger, if more event descriptions from a discourse unit were taken into account than only the ground truth mentions.

We run an additional experiment with the two-step approach in which four entity types were bundled into one entity slot. Locations, times, human and non-human participants were combined into a cumulative entity slot 
resulting in a simplified two-slot template. When using two-slot templates for both, document and sentence classification on the ECB+ 70\% CoNLL F score was reached. 
This is 3\% less than with five-slot templates. 

\section{Door-to-door implementation}
\label{sec:implementation}
In the previous sections \ref{sec:gold-data} and \ref{sec:evaluation-true}, true mentions have been used to evaluate the approach. In this section, we describe how we extract event data from raw text (system mentions) and then establish cross-document event coreference in a two-step-approach.

For extracting event data from text we use the NLP pipeline developed in the NewsReader project.\footnote{www.newsreader-project.eu}  NewsReader applies a cascade of semantic modules among which: named-entity recognition and classification (NERC), named-entity-linking (NEL), semantic role labeling (SRL), time expression detection and normalization, time anchoring, event and entity coreference. The same entity, event or time expression can be mentioned several times in a document. The above modules will interpret each occurrence separately and annotate the tokens accordingly. The output of the NLP pipeline is thus mention-based. 

From the mention-based annotation of tokens, we derive an instance-based representation of events, entities, time objects and relations between them. Our instance-based model follows the Simple Event Model (SEM, \cite{vanhage-sem}). SEM is an RDF model for presenting event-instances through URIs (Unique Resource Identifiers) with triple relations to participants and dates, which also have unique URIs. We extended SEM with the Grounded Annotation Framework (GAF, \cite{FOKKENS:2014isa}) to link each instance URI to mentions in the source documents. In Figure~\ref{fig:sem-entities}, we show the SEM representations for the entities \textit{Ka'loni Flynn} and \textit{Christopher Simpson} resulting from processing topic 45 in ECB+. Each RDF structure has a URI representing an entity as the subject and various properties, such as the \textit{rdfs:label} for the surface forms, \textit{skos:prefLabel} for the most frequent surface form and \textit{gaf:denotedBy} to link to char offsets of mentions. We can see that all mentions are offsets in different ECB+ files and there are no mentions in ECB files, which is correct.

For both persons, we see that not every mention was resolved to the same URI. For example, \textit{Christopher Simpson} has one DBpedia URI, one representation as an entity without a DBpedia match and one representation as a so-called non-entity, i.e. phrases not detected as an entity by NERC but playing an important role in an event.\footnote{We used FrameNet frame elements to decide on relevance of a participant} In the case of \textit{Ka'loni Flynn}, there is no DBpedia entry to match to and we see 3 entities and 1 non-entity. There are various reasons why the NLP modules did not match these mentions to the same entity. Such differences in URIs may also hamper the matching of events as we will see later.

%
%

\begin{figure}[ht]\tiny{
\begin{lstlisting}[basicstyle=\tiny,numbers=left]
    dbp:Christopher_Simpson>
            rdfs:label      "Christopher Simpson" , "Simpson";
            gaf:denotedBy   
            	nwr:45_5ecbplus#char=167,186 ,  nwr:45_5ecbplus#char=306,325 , 
            	nwr:45_5ecbplus#char=607,626 ,  nwr:45_5ecbplus#char=1170,1189 ,
            	nwr:45_5ecbplus#char=2516,2535 ;
	    skos:prefLabel     "Christopher Simpson" ;

    nwr:/entities/ChristopherKenyonSimpson>
            rdfs:label      "Christopher Kenyon Simpson" ;
            gaf:denotedBy   
            	nwr:45_7ecbplus#char=567,593 , nwr:45_6ecbplus#char=405,431 , 
            	nwr:45_11ecbplus#char=476,502 ,  nwr:45_8ecbplus#char=476,502 , 
            	nwr:45_2ecbplus#char=491,517 ,  nwr:45_4ecbplus#char=371,397 , 
            	nwr:45_1ecbplus#char=532,558 ,  nwr:45_12ecbplus#char=287,313 .
            skos:prefLabel     "Christopher Kenyon Simpson" .

    nwr:non-entities/purportedly+simpson
            rdfs:label         "Purportedly Simpson" ;
            gaf:denotedBy      nwr:45_5ecbplus#char=2374,2393 ;
            skos:prefLabel     "Purportedly Simpson" .
            		
    nwr:entities/KaloniFlynn>
            rdfs:label      "Ka'loni Flynn" ;
            gaf:denotedBy   
            		nwr:45_7ecbplus#char=639,652  ;
            skos:prefLabel  "Ka'loni Flynn" .
            
     nwr:entities/KaLoniMarieFlynn>
            rdfs:label      "ka'loni Marie Flynn" ;
            gaf:denotedBy   
            		nwr:45_2ecbplus#char=564,583 , nwr:45_4ecbplus#char=444,463 , 
			nwr:45_9ecbplus#char=388,407, nwr:45_3ecbplus#char=497,516 ;
            skos:prefLabel  "ka'loni Marie Flynn" .
            
    nwr:entities/KaloniFlynn>
            rdfs:label      "Ka'loni Flynn" ;
            gaf:denotedBy   
            		nwr:45_7ecbplus#char=639,652 ;
            skos:prefLabel  "Ka'loni Flynn" .
 
    nwr:non-entities/flynn>
            rdfs:label         "Flynn", 'Flynn's" ;
            gaf:denotedBy      
            		nwr:45_12ecbplus#char=584,589 , nwr:45_12ecbplus#char=733,738  , 
            		nwr:45#11ecbplus#char=778,783
            &word=w161&term=t161&sentence=8> ;
            skos:prefLabel     "Flynn"  .           
\end{lstlisting}}
\caption{Entities in SEM format with gaf:denotedBy links to mentions}
\label{fig:sem-entities}
\end{figure}

In the same way as for entities, also events are represented through unique URIs with properties as shown in Figure~\ref{fig:sem-event}. Since events are normally not stored in DBpedia and cannot be identified by their surface forms, we create meaningless unique identifiers (\textit{nwr:45\_12ecbplus\#ev10} and \textit{nwr:45\_6ecbplus\#ev16}). We further see similar properties as for the entities, such as rdfs:label, skos:prefLabel and gaf:denotedBy. We do not show the full list of mentions linked by the gaf:denotedBy property to save space but both events have mentions across various ECB+ files, implying that information has been aggregated from mentions in these files. 

Other properties for events are the class information (property \textit{a}), sem:Actor relations and a sem:hasTime relation. The class information consists of WordNet synsets \cite{wordnet}\footnote{WordNet synsets are represented here as Inter Lingual Index (ili) records: \url{www.globalwordnet.org/ili} \cite{VOSSEN-ILI:2016}. This enables us to compare events across different languages \cite{2016kbs}}, their hypernyms and the FrameNet frames associated with the events \cite{framenet}. The WordNet synsets are selected from the highest scoring senses of all the mentions according to word-sense-disambiguation (WSD). As for the actors, we listed only the entities using their URIs with their surface forms between brackets.

\begin{figure}\tiny{
\begin{lstlisting}[basicstyle=\tiny,numbers=left]
nwr:45_12ecbplus#ev10
	rdfs:label
		murder , kill , assassination , execution , Killing , 
        Shooting , slaying ;
	skos:prefLabel	murder ;
    gaf:denotedBy
		nwr:45_1ecbplus#char=1808,1815 , nwr:45_12ecbplus#char=109,115 , 
		nwr:45_5ecbplus#char=3281,3287 , nwr:45_6ecbplus#char=99,107 , 
		nwr:45_1ecbplus#char=1906,1913 , nwr:45_1ecbplus#char=5673,5686 ,
		etc... ;
	a
		ili:i28310 ,  ili:i28306 ,  ili:i28311 , ili:i34133 ,  
        ili:i36562 , ili:i35417,  ili:i34134 , ili:i34139 , 
        ili:i34130
		fn:Killing , fn:Attack ,fn:Execution , , sem:Event , ;
	sem:hasActor
		dbp:Jerome_Flynn (Flynn , Herbert Flynn , his , Ka'Loni Flynn , 
			Ka'Loni Flynn's , Ka'loni Flynn) ;
	sem:hasTime	nwr:45_6ecb#tmx2 (time:20121112 , Nov. 12) .

nwr:45_6ecbplus#ev16
	rdfs:label	charge , shooting , shoot ;
	skos:prefLabel	shoot ;
	gaf:denotedBy
		nwr:45_9ecbplus#char=640,644 , nwr:45_2ecbplus#char=633,637 , 
		nwr:45_4ecbplus#char=513,517 , nwr:45_7ecbplus#char=403,411 , 
		nwr:45_2ecbplus#char=359,366 , nwr:45_7ecbplus#char=69,77 ,
		etc... ;
	a
		ili:i106612 , ili:i25451 , ili:i25858 , ili:i25860 , 
        ili:i25976 , ili:i26598 , ili:i26600 , ili:i27206 ,
        ili:i27278 , ili:i27293 , ili:i27599 , ili:i29722 ,
        ili:i30898 , ili:i30954 , ili:i32022 , ili:i32053 ,
        ili:i33338 , ili:i34100 , ili:i34141 , ili:i35084 , 
		ili:i36049 , ili:i36050 , ili:i36591 , ili:i40503 ,
        ili:i70941 , ili:i27599 , ili:i32022 , ili:i26598 ,
        ili:i33338 , ili:i36049 , ili:i30898 , ili:i106612 ,
        ili:i27278 , ili:i26600 , ili:i25976 ,
		fn:Commerce_collect , fn:Motion , fn:Process_continue , 
		fn:Commerce_pay , fn:Killing , fn:Notification_of_charges ,  
		fn:Hit_target , fn:Shoot_projectiles , fn:Use_firearm ;
	sem:hasActor	
		dbp:Electoral_division_of_Flynn ,
		dbp:Jerome_Flynn (Flynn , Herbert Flynn , his , Ka'Loni Flynn ,
        	Ka'Loni Flynn's , Ka'loni Flynn) ,
		dbp:Oklahoma (okla , Oklahoma , Okla , Okla - man) , 
        dbp:Robb_Flynn (Ka'loni Flynn , Flynn) , 
		dbp:Fort_Smith,_Arkansas , 
        nwr:entities/ChristopherKenyonSimpson , 
		dbp:Christopher_Simpson , 
		nwr:entities/Spiroman , 
		dbp:Arkansas , 
		dbp:O._J._Simpson (Purportedly Simpson , Simpson , his) ;
    sem:hasTime	nwr:45_6ecb#tmx2 (time:2012 , 2012) .

\end{lstlisting}}
\caption{Events in SEM}
\label{fig:sem-event}
\end{figure}


Note that actors show some degree of lumping of mentions and in some cases have wrong URIs, e.g. \textit{dbp:Jerome\_Flynn (Flynn , Herbert Flynn , his , Ka'Loni Flynn , Ka'Loni Flynn's , Ka'loni Flynn)}, where references to both the murdered daughter \textit{Ka'Loni Flynn} and her father \textit{Herbert Flynn} got the same URI \textit{dbp:Jerome\_Flynn}, which is also to the wrong person. This lumping is mainly the result of wrong links coming from DBpedia spotlight \cite{mendes2011dbpedia} which gives preference to more popular entities. Wrong URIs will not harm our coreference matches as long as they are systematic, i.e. all mentions of \textit{Simpson} get matched to the same URI \textit{dbp:O.\_J.\_Simpson}. When different URIs are given, they can still be matched through surface forms (see below). Finally, we see that the events are linked to a date (sem:hasTime), which is a specific day for the first event and a year for the second event.

The above event instance representations are the result of a two-step approach that is implemented as follows:

\begin{enumerate}
\item Collect all event data for a single document and represent this as a SEM instance, giving access to all information on the action, the participants and the dates present in a document.
\item Compare the aggregated SEM representations across different documents to decide on identity. If identity is established then the SEM instance representations are merged.
\end{enumerate}

For the first step, we collect all event mentions from a single document and then collect all participants and time anchors for these mentions into a single instance representation. We first group all mentions of the same lemma and next we determine the similarity across lemmas on the basis of the WordNet similarity scores \cite{LeacockChodorow:1998} of their WordNet senses with the highest WSD score. Next, we check the Semantic Role structure created by the NewsReader pipeline to find all the roles for these mentions as well as all the time expressions to which these mentions are anchored. We thus already get aggregated instance representations with multiple surface forms across mentions in different sentences with their WordNet synsets, FrameNet frames, dates and actors.

In the second step compare these event structures across the different documents within a topic using the following heuristic:

\begin{enumerate}
\item the event actions need to match sufficiently (above a threshold) in terms of WordNet synsets or, if there are no synsets, in terms of the surface forms;
\item the time needs to match if present;
\item there must be overlap of actors in any role or the specified roles;
\end{enumerate}

For comparing actions, we check the proportion of overlap across the WordNet synsets. In case two events have no synsets associated, we use the surface forms. If the overlap is mutually above a predefined threshold, we continue, otherwise the events do not match. 

After passing the test for action similarity, we compare the overlap of the participants. We first check the URI. If the URI does not match, we check if the preferred form matches any of the surface forms. Participant matches can be required for specific semantic roles, e.g. PropBank \cite{propbank} A0, A1, A2, or the role can be ignored. At least one participant needs to match in any role or, in the former case, per specified role.

If the above test fails, there is no coreference, otherwise we continue to compare the time anchors. Time is matched either per year, month or day, where a more specific time constraint also requires the more general ones, i.e. same month also implies same year, and same day also implies the same month and year. If one event has no time anchor and the other does, there is no coreference. If both events have no time anchor, they match.

In the case of Figure \ref{fig:sem-event}, the two event structures did not get merged by our software. Their time anchors matched in terms of the year and there is an overlapping actor but none of the WordNet synsets overlap.  Nevertheless, each of the event instances shows already quite some lumping of other mentions across the documents, as indicated by the mentions in different files in topic 45.

\section{Experiments with coreference on system mentions}
\label{sec:evaluation}
We evaluate the NewsReader system on system mentions on the same ECB+ data set. We compare the NewsReader results with Yang et al. (2015) \cite{YangCF15}, who report the best results for event coreference resolution system mentions for ECB+ and who also compare their results to other systems that have so far only been tested on ECB and not on ECB+. Yang et al use a distance-dependent Chinese Restaurant Process (DDCRP \cite{blei2011distance}), which is an infinite clustering model that can account for data dependencies. They define a hierarchical variant (HDDCRP) in which they first cluster event mentions and data within a document and next cluster the within document clusters across documents. Their hierarchical strategy is similar to our approach using event components, in the sense that event data can be scattered over multiple sentences in a document and needs to be gathered first. Our approach differs from theirs in that we use a semantic representation to capture all event properties and do a logical comparison, while Yang et al. used machine learning methods (both unsupervised clustering and supervised mention based comparison). Yang et al. also report on a lemma-baseline as proposed by Cybulska and Vossen (2014) \cite{CYBULSKA+VOSSEN:2014}, where all event mentions with the same lemma within and across documents are simply joined in a single coreference set.

Yang et al. test their system on topics 24-43 while they used topics 1-20 as training data and topics 21-23 as the development set. They do not report on topics 44 and 45. To compare our results with theirs, we also used topics 24-43 for testing. In Table~\ref{tab:cross-doc-event-coref}, we show Yang's lemma baseline (LEMMA), Yang's best results (HDDCRP), and the results for NewsReader (NWR). The  NewsReader systems are not trained on ECB+ data and use logical comparison of event data. We tested the following variants of the NewsReader system, where systems starting with \textbf{NWR-X} (out-of-the-box), \textbf{NWR-T} (with TimeEval2013 training for event detection) or \textbf{NWR-G} (with true mentions of events). The remainder of the code expresses the time filter (Y=year, M=month, D=Day, N=none), the participant filter (A=any role, A1=PropBank A1, N=none) and the action filter (c10,30,50,70=concept overlap, p10,30,50,70=phrase overlap):

{\small
\begin{description}
\item[NWR-X-YAc30p30] NewsReader out-of-the-box, matching year (Y), any participant (A), concept overlap of 30\% (c30), phrase overlap of 30\% (p30).
\item[NWR-T-YAc30p30] NewsReader with event extraction using CRF trained on TemEval-2013 training data (T), matching year (Y), any participant (A), concept overlap of 30\% (c30), phrase overlap of 30\% (p30).
\item[NWR-G-YAc30p30] Newsreader with Gold event data (G), matching year (Y), any participant (A), concept overlap of 30\% (c30), phrase overlap of 30\% (p30).
\item[NWR-G-MAc30p30] Newsreader with Gold event data (G), matching month (M), any participant (A), concept overlap of 30\% (c30), phrase overlap of 30\% (p30).
\item[NWR-G-DAc30p30] Newsreader with Gold event data (G), matching day (D), any participant (A), concept overlap of 30\% (c30), phrase overlap of 30\% (p30).
\item[NWR-G-YAc10p10] Newsreader with Gold event data (G), matching year (Y), any participant (A), concept overlap of 10\% (c10), phrase overlap of 10\% (p10).
\item[NWR-G-YAc50p70] Newsreader with Gold event data (G), matching year (Y), any participant (A), concept overlap of 50\% (c50), phrase overlap of 50\% (p50).
\item[NWR-G-YAc70p70] Newsreader with Gold event data (G), matching year (Y), any participant (A), concept overlap of 70\% (c70), phrase overlap of 70\% (p70).
\item[NWR-G-YNc30p30] Newsreader with Gold event data (G), matching year (Y), no participant (N), concept overlap of 30\% (c30), phrase overlap of 30\% (p30).
\item[NWR-G-YA1c30p30] Newsreader with Gold event data (G), matching year (Y), A1 participant (A), concept overlap of 30\% (c30), phrase overlap of 30\% (p30).
\item[NWR-G-NAc30p30] Newsreader with Gold event data (G), no time (N), any participant (A), concept overlap of 30\% (c30), phrase overlap of 30\% (p30).
\end{description}
}

\begin{table}
{\scriptsize 
  \centering
  \caption{Reference results macro averaged over ECB+ corpus as reported by Yang et al. \cite{YangCF15} for state-of-the-art machine learning systems as compared to various NewsReader based systems. All NewsReader systems start with NWR-X (out-of-the-box), NWR-T (with TimeEval2013 training for event detection) or NWR-G (with true mentions of events). The remainder of the code expresses the time filter, the participant filter and the action filter. More explanation is given in the text.}   \label{tab:cross-doc-event-coref}    \begin{tabular}{lrrrrrrrrrrr}
    \hline
    ECB+  & \multicolumn{3}{c}{MUC} & \multicolumn{3}{c}{BCUB} & \multicolumn{3}{c}{CEAFe} & CoNLL & Mention\\
    \hline
    Topics 24-43 & R     & P     & F$_1$     & R     & P     & F$_1$     & R     & P     & F$_1$     & F$_1$  & F$_1$\\
    \hline
    LEMMA & 55.4 & 75.10 & 63.80 & 39.60 & 71.70 & 51  & 61.10 & 36.20 & 45.50 & 53.40 & 95 \\
    \textbf{HDDCRP} & 67.10 & 80.30 & 73.10 & 40.60 & 73.10 & 53.50 & 68.90 & 38.60 & 49.50 & \textbf{58.70} & 95\\
    \hline
    NWR-X-YAc30p30 & 44.85 & 50.16 & 47.35 & 46.88 & 45.3  & 46.08 & 47.45 & 34.89 & 40.22 & 44.55 & 67.99\\
    NWR-T-YAc30p30 & 48.99 & 58.5  & 53.33 & 45.37 & 55.48 & 49.92 & 41.37 & 45.56 & 43.36 & 48.87 & 75.03\\
    NWR-G-YAc30p30 & 64.12 & 72.03 & 67.85 & 65.21 & 74.89 & 69.72 & 66.35 & 57.39 & 61.55 & 66.37 & 99.84 \\
    NWR-G-MAc30p30 & 64.12 & 72.03 & 67.85 & 65.21 & 74.89 & 69.72 & 66.35 & 57.39 & 61.55 & 66.37 & 99.84 \\
    NWR-G-DAc30p30 & 62.12 & 70.99 & 66.26 & 61.93 & 75.69 & 68.12 & 66.57 & 56.52 & 61.14 & 65.17 & 99.84\\
    \hline
    NWR-G-YAc10p10 & 64.81 & 70.6  & 67.58 & 65.57 & 72.84 & 69.02 & 63.75 & 57.1  & 60.24 & 65.61 & 99.84\\
    NWR-G-YAc50p50 & 63.49 & 72.55 & 67.72 & 64.63 & 75.84 & 69.79 & 67.48 & 57.29 & 61.97 & 66.49 & 99.84\\
    NWR-G-YAc70p70 & 62.61 & \textbf{72.81} & 67.33 & 63.8  & 76.92 & 69.75 & 67.9  & 56.61 & 61.74 & 66.27 & 99.84\\    
    \hline
    NWR-G-YNc30p30 & \textbf{77.4} & 69.68 & 73.34 & \textbf{72.92} & 64.24 & 68.31 & 54.99 & \textbf{65.39} & 59.74 & \textbf{67.13} & 99.84\\
    NWR-G-YA1c30p30 & 52.31 & 71.27 & 60.34 & 58    & \textbf{80.27} & 67.34 & \textbf{69.89} & 50.67 & 58.75 & 62.14 & 99.84\\
    NWR-G-NAc30p30 & 64.12 & 72.03 & 67.85 & 65.21 & 74.89 & 69.72 & 66.35 & 57.39 & 61.55 & 66.37 & 99.84\\
    \hline
    \end{tabular}}
\end{table}%

We first compare the NewsReader out-of-the-box system (NWR-X-YAc30p3) with Yang's results (LEMMA baseline and HDDCRP). The NewsReader system uses the following matching settings: year, a single participant in any role and 30\% of the event concepts or, if not present, the event surface forms. We see that both Yang's \textit{HDDCRP} and the lemma baseline outperform NWR-X-YAc30p3 by 14 and 9 points respectively in CoNLL F$_1$ score \cite{Phradan-etal-2011}.
However, Yang et al. report that their system at first had an out-of-the-box accuracy for event detection of 56\%. They therefore trained a separate Conditional Random Field (CRF) event detection system with event annotations of the first 20 topics (about half of the data set). They report an accuracy for this classifier of 95\% on event detection and they used it as input for both the LEMMA baseline and HDDCRP. For comparison, the NewsReader system has an out-of-the-box accuracy of 67.99\%, where events are detected by the MATE tool \cite{Bjorkelund:2010} which is trained on PropBank data. Clearly, what events have been annotated and how they were annotated has a big impact on the results. To measure this impact on the actual event-coreference, we created two  other variants of the NewsReader system: 1) NWR-T replaces the MATE event detection by a CRF classifier trained with SemEval 2013 - TempEval3 gold data \cite{uzzamansemeval2013} and 2) NWR-G uses the true mentions of the ECB+ annotation as the events (Gold). The event detection accuracy of NWR-T is  75.03\% and the accuracy for NWR-G is 99.84\%.\footnote{The reason that NWR-G is not 100\% is because the NewsReader system could not process one of the evaluation files due to formatting problems.}  

In the last column of Table~\ref{tab:cross-doc-event-coref}, we list the F$_1$ measures for the detection of event mentions by each variant system. It clearly shows that the differences in coreference results across systems are mainly due to the performance on the event detection. The NWR-G variants for example outperform HDDCRP by almost 10 points, while scoring only 5 points higher in event detection. NWR-T-YAc30p30 performs 7 points higher in event detection and 4 points higher in event coreference than NWR-X-YAc30p30. The NWR-X-YAc30p30, NWR-T-YAc30p30, NWR-G-YAc30p30 only differ in the extraction of the events with accuracies of 68\%, 75\% and 100\% respectively. The other parameters for time, participant and action match are the same. Note that all NewsReader systems apply logical comparison and are not trained on the ECB+ data set. We thus can expect their performance to be relatively stable across data sets, whereas Yang et al's system is expected to perform significantly lower when applied to out-of-domain data.

Next, Tabe \ref{tab:cross-doc-event-coref} lists the applied different variants of the NewsReader system using the true mentions of events (NWR-G) to see the impact on event coreference given the perfect event detection. Varying the settings for matching using the true mentions for events, shows only small differences. When we make the time more strict (year (Y), month (M), day (D)), we see that the month is as discriminative as the year but CoNLL F$_1$ is 1 point lower when the time needs to match at the level of the day. Next, we varied the threshold for overlapping concepts and surface forms (10\%, 30\%, 50\%, 70\%). We can see that precision for MUC and BCUB are higher when the thresholds are higher. This is in line with our expectation. For CEAFe, we see the same trend for recall. The last rows of Table~\ref{tab:cross-doc-event-coref} show the impact of the participants. In case of NWR-G-YNc30p30, no matching participant is required, for NWR-G-YA1c30p30 the PropBank A1 role should be identical, and for NWR-G-NAc30p30 we match any participant but we dropped the time constraint. Remarkably, dropping the participant match constraint gives best results, while requiring a matching A1 participant gives highest precision scores for BCUB and recall for CEAFe.

Overall, we can conclude that there is a slight tendency for more strict parameters to increase the precision but that we always loose more recall with slightly lower F$_1$ scores as a result. It is also clear that the event detection itself is the most important factor for improving event coreference. 

Note that the best performance obtained with the true mentions of the events (67.13), scores only 6 points below our bag-of-events approach (73) using true mentions for all event components and it outperforms the lemma baseline on true mentions (63) reported in section \ref{sec:evaluation-true}. Obviously, the NewsReader approach uses more data from the document than the annotated data (1.8 sentences) but on the other hand it also introduces more noise. We can expect that improving the linking of participants to entity mentions and improving the event detection is likely to bring the out-of-the-box system closer to 73 F$_1$ scores.


\section{Discussion}
\label{sec:discussion}
The results have shown that cross-document event coreference is a hard task that is not completely solved despite the progress. We demonstrated that event components are critical but that they need to be collected from the complete document. Nevertheless, we have also seen that event detection as such is a major factor for the current performance starting from raw text.

With respect to the granularity of the matching of the components, we have used various ways to abstract from surface forms:

\begin{itemize}
\item different forms are matched to the same URI assigned by DBpedia Spotlight, created from surface forms or through nominal coreference;
\item time expressions are normalised to the ISO dates;
\item event mentions with different forms are matched through WordNet similarity and their word-senses;
\end{itemize}

Furthermore, we can parameterize the matching by setting loose or strict constraints:

\begin{itemize}
\item dates can be mapped by year or month instead of day;
\item more or less participants to be shared, with or without their roles;
\item degree of overlap of concepts and the range of concepts above a word-sense-disambiguation (WSD) threshold; 
\end{itemize}

We have seen that making these constraints more tight results in more precision and lower recall. Making them more loose has the opposite effect.

There are still mentions that are not mapped to the same instances, e.g. "his girlfriend", "the daughter" to "Ka'loni Flynn", and there are many missed URIs as well as wrong URIs assigned. The quality of modules such as NERC, NED and WSD is crucial in this respect. Furthermore, time-anchors are very sparse and difficult to infer from the text as such.\footnote{We left out the document creation time as a baseline time-anchor because it may interfer with the task since the articles on each seminal event were published on different dates.}

However, it is also the case that the ECB+ database is still too restricted to measure the true contribution of the component-based approach. We have seen that matching years instead of months makes no difference due to the fact that the events can already be distinguished by the year. Adding more seminal events to create more referential ambiguity for similar events around similar periods in time will require more precise analysis and component matching. 

Defining the granularity of event descriptions provides an interesting view on event-coreference. How far can we go to lump together event data? In a way, we could lump all events that make up a story or a topic together and define a period of time in which the topic or story takes place with all the involved participants. This does not necessarily violate the idea of event-coreference since peoples' intuitions on decomposing events to smaller units are also not clear-cut. Obviously, at some point lumping of event data generates unclarity of scope relations between events and participants, such as more than one person murdering the same or different persons, or even semantic anomalies such as or the same person being at different places at the same time. This is where event coreference could set a hard border but this also means that annotation and evaluation of data sets may need to be different, e.g. assigning not only event-coreference relations but also subevent and topical relations.

A final aspect that still needs to be investigated is variation. Even though ECB+ has documents from various sources, events that are annotated as coreferential have mostly the same lemma and most events have no coreference relation at all (90\% of all mentions). Annotation of event coreference is not an easy task and annotators tend to be conservative. More variation in reference to events is also expected to put higher demands on a semantic approach rather than approaches that are trained on mentions only.

\section{Conclusion}
\label{sec:conclusion}
We described a new method to detect event descriptions in text and to model semantics of events, their components and event coreference using RDF representations. The proposed heuristic outperforms the state-of-the-art system when assuming equal quality in event detection. Our approach collects the component information from the complete document rather than from the local context of the event mentions that are compared. We have shown that event components play a role in obtaining precision and recall and that their matching needs to be adapted to the granularity of the task.

In future work, we want to investigate event-coreference in relation to topical structures and storylines. We believe that this also helps creating annotated data in which more variation and referential ambiguity is reflected. This will make both the annotation of data and the task more natural. Such data sets provide additional possibilities and challenges to obtain more precise temporal ordering of events for event coreference.

\subsubsection*{Acknowledgments.} The NewsReader project was co-funded by the European Union as project number: 316404, FP7 Work Programme Call FP7-ICT-2011-8 Ð Objective Cooperation Research theme "Information and Communication Technologies", challenge 4.4 - Area Intelligent Information Management. 

\bibliographystyle{splncs03}
\bibliography{bibliography}

%
%
%
%
%
%
%

\end{document}